\documentclass[lettersize,journal]{IEEEtran}
\usepackage{amsmath,amsfonts}
\usepackage{algorithmic}
\usepackage{array}
\usepackage[caption=false,font=normalsize,labelfont=sf,textfont=sf]{subfig}
\usepackage{textcomp}
\usepackage{stfloats}
\usepackage{url}
\usepackage{verbatim}
\usepackage{graphicx}
\usepackage{xcolor}
\def\BibTeX{{\rm B\kern-.05em{\sc i\kern-.025em b}\kern-.08em
    T\kern-.1667em\lower.7ex\hbox{E}\kern-.125emX}}
\usepackage{balance}
\usepackage[colorlinks=true,allcolors=blue]{hyperref}
\usepackage{cite}
\usepackage[absolute,overlay]{textpos}
\usepackage{everypage}

\newcommand{\distroA}[0]{
	\begin{textblock*}{25cm}(2.75cm,26.75cm) 
		Distribution Statement A: Approved for Public Release; Distribution is Unlimited. PA AFRL-2024-2772.
	\end{textblock*}
}
\AddEverypageHook{\distroA}

\begin{document}
	\distroA
	
	\title{Iterative Learning Control of Fast, Nonlinear, Oscillatory Dynamics}
\author{John~W.~Brooks,
	and~Christine~M.~Greve
	\thanks{J. Brooks is with the U.S. Naval Research Laboratory, Washington, D.C., USA e-mail: john.w.brooks197.civ@us.navy.mil.}
	\thanks{C. Greve is with the  Air Force Research Laboratory, Edwards AFB, California 93524, USA}}

\markboth{Preprint}%
{}


\maketitle

\begin{abstract}

The sudden onset of deleterious and oscillatory dynamics (often called instabilities) is a known challenge in many fluid, plasma, and aerospace systems.  These dynamics are difficult to address because they are nonlinear, chaotic, and are often too fast for active control schemes.  In this work, we develop an alternative active controls system using an iterative, trajectory-optimization and parameter-tuning approach based on Iterative Learning Control (ILC), Time-Lagged Phase Portraits (TLPP) and Gaussian Process Regression (GPR).  The novelty of this approach is that it can control a system's dynamics despite the controller being much slower than the dynamics.  We demonstrate this controller on the Lorenz system of equations where it iteratively adjusts (tunes) the system's input parameters to successfully reproduce a desired oscillatory trajectory or state.  
Additionally, we investigate the system's dynamical sensitivity to its control parameters,  identify continuous and bounded regions of desired dynamical trajectories, and demonstrate that the controller is robust to missing information and uncontrollable parameters as long as certain requirements are met.  
The controller presented in this work provides a framework for low-speed control for a variety of fast, nonlinear systems that may aid in instability suppression and mitigation.

\end{abstract}


\section{Introduction}


	The sudden onset of deleterious and oscillatory dynamics (often called instabilities) is a known challenge in many fluid and plasma systems.  
	Examples within the aerospace community include: air-breathing and rocket combustion instabilities \cite{oyediran1995, biggs2009, zhao2018,liu2020}, Hall-thruster plasma instabilities \cite{choueiri2001}, aeroelastic instabilities (i.e. flutter) \cite{jonsson2019}, hypersonic boundary layer instabilities \cite{fedorov2011}, and two-phase instabilities in closed-loop heat pipes \cite{khandekar2004}. 
	These dynamics are difficult to model, predict, and ameliorate because they are nonlinear, chaotic, have dynamics that span multiple time and spatial scales, are often poorly understood, and are often too fast for modern controllers and actuators.  
	They are problematic because they result in performance loss, decreased lifetime, and system failures (sometimes catastrophic), and they are often not discovered until qualification or flight.  This can lead to project delays, cost overruns, and cancellations.   
	A notable example of this was with the National Aerospace Plane Program (NASP), which was canceled, in no small part, because the design team was unable to confidently predict the location of boundary layer transitions from laminar to turbulent flow~\cite{shea1992}.
	
	To mitigate these undesired dynamics, the community commonly employs one of the following four approaches.
	The first is ``performance derating'' where designers lower the system's rated performance to diminish the deleterious dynamics.  Performance derating is often under-reported. 
	The second method is ``parameter mapping'' where discrete, safe-operating points in the input parameter space are identified for flight.   A common example is creating IVB maps in Hall thrusters which are made by adjusting discharge voltage (V) and the magnetic coil strength (B) until oscillations in the discharge current (I) are at reasonable levels\cite{kamhawi2016}.  A common problem with this approach is that these points are often developed in simulation or in a laboratory environment that do not fully reproduce flight conditions~\cite{byers2009}, and therefore the safe operating points may not manifest during flight.  We refer to this phenomena as facility-to-flight reproducibility issues.  
	The third method is passive control where the system is redesigned so that the underlying dynamics responsible for the instability are decoupled.   A notable example of this is the addition of acoustic baffles in the F1 engine\cite{bostwick1968}, the lower stage engine of the Saturn V rocket.  These redesigns often involve significant testing, which adds cost, complexity, and development time to the project. 
	The final method is active control where a controller (open or closed-loop) uses actuators to suppress or otherwise modify the dynamics.  	A modern example is tangentially injecting air into hypersonic boundary layers to suppress oscillations or control the location of the transition~\cite{joslin2012, yang2022}.  However, active control is often not viable because fluid and plasma dynamics have characteristic frequencies in the kHz to MHz, and the controller and actuator often cannot operate this fast~\cite{concezzi2012, brooks2020}.  These hardware limitations effectively place an upper frequency limit on active, real-time, control of these dynamics.
	
	Iterative Learning Control (ILC)~\cite{bristow2006, ji2018, zhang2019, vemula2022} is a controls framework that uses intelligent parameter tuning to bypasses this frequency limit by using slow (not real-time) feedback and low-speed actuation.  ILC works by iteratively testing a system's response to preprogrammed, quasi-static control inputs and using reinforcement learning to learn the inputs that achieve a desired behavior.  
	Because ILC does not require fast (i.e. real-time) processing and actuation to operate, it is therefore capable of controlling high-speed dynamics with low-speed control, assuming the system is sufficiently sensitive to the low-speed actuation.  	
	ILC has historically been used in robotics to teach an assembly-line robots sequences of movements (a trajectory) in order to perform a task~\cite{norrlof2002}.  	
	ILC has recently begun transitioning to high-speed fluid/plasma applications; examples include controlling aerospace flutter~\cite{xu2015} and the rapid collapse of plasma confinement (disruptions) in tokamaks~\cite{ravensbergen2017}.  
	A traditional shortcoming of ILC is that it struggles to control non-repetitive dynamics (e.g. non-linear, chaotic, noisy).  In this work, we address these shortcomings with several advances in Bayesian optimization and system identification, specifically Gaussian Process Regression and Time-Lagged Phase Portraits.  
	
	Gaussian Process Regression~\cite{rasmussen2005, ghahramani2015, wang2023} (GPR) is a non-parametric, Bayesian-optimization technique common to the machine learning community, and because of its statistical nature, it is well suited to optimizing non-repetitive dynamics.  GPR works by building a Gaussian Process (GP) model of an objective function, predicting parameters that minimize the GP model and its uncertainty, and incorporating new measurements to improve the GP model (i.e. reinforcement learning) after each iterative test.  It is this focus on statistical uncertainty that allows GPR to more robustly converge as compared with traditional gradient descent methods, which struggle to minimize functions that are noisy or otherwise non-repetitive.  Notable examples include using GPR to learn nonlinear system dynamics in quadcopters~\cite{berkenkamp2015} and remote control race cars~\cite{hewing2020}, and also speeding-up computational fluid dynamic simulations~\cite{umetani2018}.  GPR is known for being computationally expensive (i.e. slow), making it unsuitable for many real-time application but not too slow for use with ILC.  An example of GPR and ILC pairing is in a recent work that compensates for nonlinear disturbances affecting aircraft trajectories~\cite{buelta2021}.  
	
	Time-Lagged Phase-Portraits (TLPP) are a type of time-lagged, trajectory representation of measurements (sometimes called time delay embedding) that are well suited for isolating non-repetitive (non-linear and chaotic) dynamics\cite{sugihara1990, greve2019,george2021}. 	In practice, the TLPP is the conversion of a measurement of a dynamical system into a higher dimensional phase-space using time-lagged signals of the same measurement, which is related to Koopman operator theory~\cite{haggerty2023}.   If a system is sufficiently coupled~\cite{takens1981} and if the correct TLPP hyperparameters (i.e. time lag) are used, high-dimensional manifolds (shapes) comprised of repeated (both periodic and non-periodic) dynamical trajectories appear in the resulting TLPP, which provides a unique representation (i.e. a ``fingerprint'') of the nonlinear dynamics.  
	The utility of TLPP in this work is that the controller is provided a reference or ``desired'' fingerprint and then adjusts its control parameters until the present fingerprint converges on (looks like) the reference fingerprint.  
	For context, a Fourier spectrum also provides a unique fingerprint of dynamics in the form of the dominant wave amplitudes and their associated frequencies.  A notable downside to the Fourier spectrum struggles with isolating non-periodic dynamics (see App.~\ref{app:f_lorenz}), which is why we use TLPP here.
	Time-embedding was popularized by Sugihara to establish causality between state variables in ecosystem dynamics\cite{sugihara1990,sugihara2012}.  Other recent uses of TLPP include reconstructing Hall thruster dynamics\cite{eckhardt2019}, quantifying unknown parameters in physics-based models\cite{greve2019}, and denoising signals\cite{araki2021}.  TLPP's have also been used in several control systems, including mode control of chaos in a reaction-diffusion system~\cite{triandaf1997} and control of inertial dynamics in soft robots~\cite{haggerty2023}.
	
	In this work, we develop an ILC system that uses GPR to learn the system's input parameter space to reproduce the desired dynamics, while avoiding undesired dynamics.   ILC is an advantageous framework because it can control dynamics that are traditionally considered too fast for real-time control.  Additionally, the TLPP and GPR both help overcome ILC's historical weakness at controlling non-repetitive (nonlinear, chaotic, noisy) dynamics. 
	The outline of this work is as follows.  Section.~\ref{sec:Lorenz} introduces the Lorenz system, a toy model with relevant dynamics that we use as a test-bed to demonstrate our control system.   Section~\ref{sec:architecture} provides an explanation of the control system and its core algorithms.  Finally, Section.~\ref{sec:results} applies this controller to the Lorenz system for three cases: i) a single-parameter control test of its input parameters to illustrate parameter sensitivity, ii) a two-parameter control test to demonstrate multi-parameter control and to identify bounded, continuous regions of acceptable dynamics, and iii) a  test to demonstrate that the controller is robust to parameters that are outside of its direct control (e.g. thermal transients), which highlights the importance of identifying and measuring dynamically sensitive parameters.

\section{Methods: The Lorenz system \label{sec:Lorenz}}
	
\begin{figure}
		\includegraphics[]{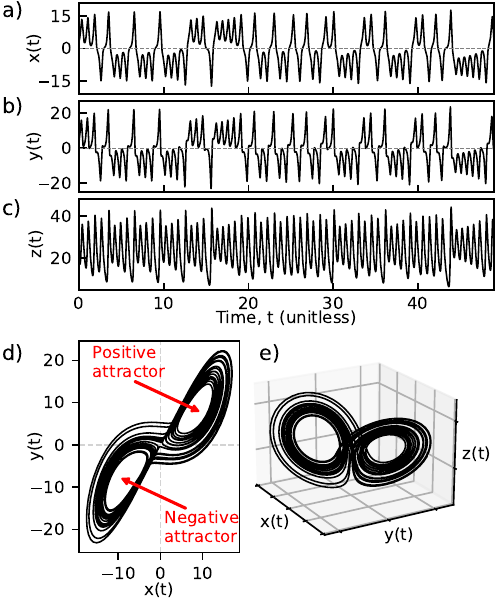}
		\caption{\label{fig:Lorenz} The Lorenz system solved at $\sigma=10$, $\rho=28$, and $\beta=8/3$.  Subfigures a, b, and c show the time evolution of $x(t)$, $y(t)$, and $z(t)$, respectively, and d and e show two and three-dimensional phase space representations, respectively.   }
	\end{figure}

	
	To demonstrate our control system in this work, we test it on a toy problem: the Lorenz system.  
	The Lorenz system \cite{Lorenz1963,sparrow1982} was originally developed as a simplified, chaotic, weather model and is comprised of three, nonlinear, coupled ordinary-differential-equations,

	\begin{subequations} \label{eq:Lorenz}
		\begin{align}
			\frac{\partial x}{\partial t}& = \sigma \left( y - x\right), \\
			\frac{\partial y}{\partial t}& = x \left( \rho - z\right) - y, \\
			\frac{\partial z}{\partial t}& = x y - \beta z 
		\end{align}
	\end{subequations} 
	
	\noindent with three state variables $\left\{ x(t) \text{, } y(t)\text{, } z(t) \right\}$ and three constant, positive parameters $\left\{ \sigma \text{, } \rho \text{, } \beta \right\}$.  The system's dynamics are strongly dependent on these parameters, and therefore in this work, we treat them as adjustable, control parameters to produce desired dynamics (i.e. control the system).  
	Figure~\ref{fig:Lorenz} shows the classic example solution of the Lorentz system solved at $\sigma=10$, $\rho=28$, and $\beta=8/3$.  Figures~\ref{fig:Lorenz}a and~\ref{fig:Lorenz}b show that both $x(t)$ and $y(t)$ oscillate around two attractors (one positive and one negative), which when visualized in phase-space (Figures~\ref{fig:Lorenz}d and~\ref{fig:Lorenz}e), provide the Lorenz system's characteristic ``butterfly'' appearance. 
	
	The Lorenz system is a particularly relevant toy problem for many aerospace applications for several reasons.  
	i) It is characterized by wave dynamics including periodic and non-periodic oscillations.
	ii) It has non-linear terms  ($xz$ and $xy$) in its governing equations, which result in non-linear features in its time evolution (e.g. the two attractors).
	iii)  These non-linear terms also result in a chaotic system whose time-series evolution is strongly dependent on initial conditions.  An example of its chaotic behavior is its seemingly random transitions (bifurcations) between its attractors.  Similar to real-life systems, both points ii and iii make accurate prediction difficult.
	iv) Its dynamics and its stability are strongly dependent on its three parameters~\cite{sparrow1982}, meaning its dynamics and stability are controllable.
	To better contextualize, Hall thrusters~\cite{boeuf2017}, a type of in-space plasma-propulsion thruster, experience plasma instabilities subject to the same four characteristics as above.  Specifically, their dynamics: 
	i) consist of coupled oscillations between the momentum of its propellant (neutral and ionized) and electromagnetic fields~\cite{choueiri2001}, ii) exhibit non-linear coupling between its state variables~\cite{barral2010}, iii) transition (``hop'') unexpectedly between instabilities\cite{hara2014,brooks2023}, and iv) are strongly dependent on multiple parameters\cite{jorns2022}, including: power supply settings, mass flow rates, geometries, temperatures, pressures, and more.  

	In this work, we use Python's  \emph{solve$\_$ivp} function within the \emph{scipy} library to solve the Lorenz system with a RK45 (Runge-Kutta) solver and a dynamic time step.  It outputs the solution at a fixed time step output, $dt=0.01$, with $2N$ points where $N=10^5$.  For each iteration, the controller or operator specifies scalar values for $\beta$, $\rho$, and $\sigma$.  Unless otherwise specified, these parameters are equal to $\left\{ \sigma \text{, } \rho \text{, } \beta \right\} = \left\{ 10 \text{, } 28 \text{, } 8/3 \right\}$.     	When solved as part of the control loop, the final time step of the previous iteration is used as the initial conditions for the next iteration.  When no previous iteration is available, we use the initial conditions: $\left\{ x(0) \text{, } y(0)\text{, } z(0) \right\}$ = $\{0.1\text{, }0.2\text{, }0.3\}$.  To ensure convergence when changing parameters (see discussion below), the solver discards the first $N$ points from each simulation, and the controller performs its analysis on the remaining $N$ points.
	
\begin{figure}
		\includegraphics[]{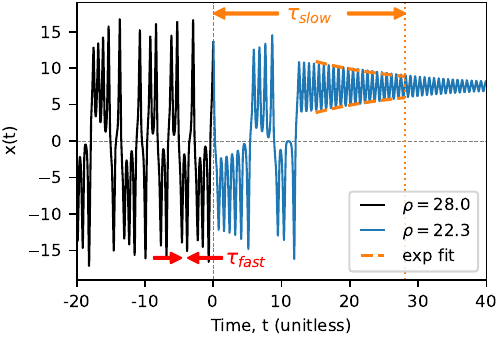}
		\caption{\label{fig:Lorenz_time_scales} We change the Lorenz system's parameter, $\rho$, at time $t=0$, and the dynamics change and converge on a solution after a finite amount of time, $\tau_{slow}$.  The dynamics of the Lorenz system occur at $\tau_{fast}$ which is over an order of magnitude smaller than $\tau_{slow}$ in this example.  }
	\end{figure}

	Figure~\ref{fig:Lorenz_time_scales} shows an example of the Lorenz system's dynamical sensitivity to its parameters and additionally shows that the Lorenz system has two distinct time scales: fast ($\tau_{fast}$) and slow ($\tau_{slow}$).  At $t<0$, the Lorenz system is solved at its default parameters.  At $t=0$, we change $\rho$ from $\rho=28.0$ to $\rho=22.3$, and the solver continues.  At $t\approx11$, the system stops bifurcating between the two attractors and commits to the positive attractor, and the oscillation exponentially decays with an e-folding time of about 17.  The total time from the parameter change to approximate convergence is around  $\tau_{slow}\approx 28$, and this response time to changing parameters how we define $\tau_{slow}$.  	
	In contrast, $\tau_{fast}$ is the period of the system's oscillations, and this typically sets an upper limit for traditional, real-time, active control.  For the Lorenz system solved at these parameters, this is $\tau_{fast} = 0.76 \ll \tau_{slow}$ (see App.~\ref{app:f_lorenz} for details).  While $\tau_{slow}$ is highly variable and dependent on many factors, it is consistently much larger than $\tau_{fast}$.  
	In summary, i) changing $\rho$ resulted in a notable change in dynamics to the system, and ii) this response occurred at the time scale, $\tau_{slow} \gg \tau_{fast}$.  Therefore, we propose that a controller designed for parameter tuning at time scales around $\tau_{slow}$ (or slower) is an effective alternative for controlling a system's dynamics.
	
\section{Methods: Control architecture \label{sec:architecture}}

	\subsection{Overview}

\begin{figure}
			\includegraphics[]{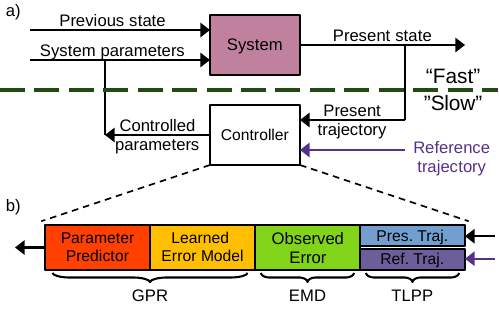}
			\caption{\label{fig:control_architecture} Block diagram of the a) control loop and b) controller.  The system naturally evolves at a ``fast'' speed, and the controller, which cannot operate that fast, instead operates (``closes the loop'') at a ``slow'' speed by controlling (tuning) the system's parameters.  
			}
		\end{figure}

		Figure~\ref{fig:control_architecture}a shows an overview of the control architecture developed in this work, which is based on an Iterative Learning Control framework and previous work~\cite{greve2019}.  It consists of the four traditional control-loop components: the system being controlled (the Lorenz attractor), measurements of the system's present state (discretized and noiseless time-series measurement of $x(t)$ only), the controller (discussed below), and the actuators (changing one or more of the three control parameters).  In this example, the loop's cycle time is set by the sum of the long measurement time ($t=2Ndt=2000$), the computational time of the controller (assumed instant), and the time required to update the actuators (assumed instant).  The result is that the controller's cycle time ($t_{loop}=2000$) is much slower than the Lorenz system's oscillations ($\tau_{Lorenz}=0.76$).  
		
		The controller is diagrammed in Figure~\ref{fig:control_architecture}b.  At its most fundamental level, the controller is composed of two elements: an objective function and an optimizer.  
		In this work, the objective function is the error between the present dynamical trajectory (a unique ``fingerprint'' of the dynamics) and a reference (desired) trajectory, and when minimized, the system produces dynamics nearly identical to the reference trajectory.  
		The trajectories are the generated Time-Lagged Phase Portrait (Sec.~\ref{subsec:tlpp}) representations, and the observed error between trajectories is calculated by the Earth Mover's Distance (Sec.~\ref{subsec:emd}). The optimizer is likewise composed of two components: a model and predictor.
		The model is a learned error model of the EMD-TLPP objective function in parameter space ($\sigma \text{, } \rho \text{, } \beta$), and the predictor uses the model to predict the next set of parameters that will minimize the objective function while reducing uncertainty in the model.  When the loop is closed, the model incorporates each new measurement, and therefore it becomes better at minimizing the objective function with each iteration.  In this work, Gaussian Process Regression (Sec.~\ref{subsec:bo}) composes both steps of the optimizer.
		Below, we describe each of these three algorithms in the context of their application to the Lorenz system.

	\subsection{Time-lag phase-portrait (TLPP) \label{subsec:tlpp}}

		
		To create a TLPP trajectory, the first step is to acquire a measurement of system's dynamics over many cycles.  In this work, we measure $x(t)$, which has the 1D discretized form, $\{x_1,\, x_2,\, \ldots,\, x_N\}$, with $N$ points where $N dt \gg \tau_{fast}$ and with no simulated measurement noise.  The second step is to choose a single time delay, $\tau_1=n_1 dt$, where $n_1$ is the integer index associated with $\tau_1$, to create a time-lagged representation of the measurement: $\mathbf{x}(t-\tau_1)$.   For this work, we use $\tau_1 = 0.17$, which is discussed in App.~\ref{app:tau}.  The third step is to combine (embed) the two discretized time-series into a 2D set of coordinate-pairs,
		
		\begin{equation} \label{eq:tlpp_2d}
			\begin{split}
				\{&(x_1,\, x_{1+n_1}), \\&(x_2,\, x_{2+n_1}),\\ & \qquad  \vdots\\ &(x_{N-n_1},\, x_{N})\}.
			\end{split}
		\end{equation}  	
		
		\noindent  In this work, we restrict our TLPPs to 2D because i) it is simpler to conceptualize, ii) it is computationally faster than higher dimensional TLPPs, and iii) 2D is sufficient to adequately capture the system's dynamics (see App.~\ref{app:tau}).  In general, however, we are not limited to 2D and could include additional $\tau$ values ($\tau_2$, $\tau_3$, \ldots) and other measurements of the system ($y(t)$, $z(t)$), each of which can be additionally time-lagged (e.g. $z(t-\tau_2)$).  
		
\begin{figure}
			\includegraphics[]{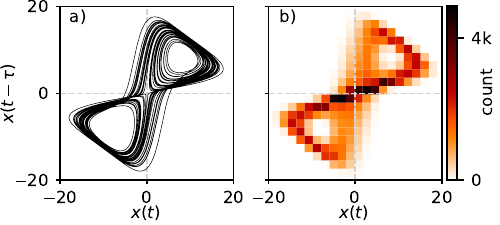}
			\caption{\label{fig:Lorenz_TLPP} The 2D TLPP of the Lorenz system: a) unbinned, and b) binned.  
				Only the first 5000 points are shown in a) for clarity of the trajectories. 
			}
		\end{figure}

		Figure~\ref{fig:Lorenz_TLPP}a provides an example of the 2D TLPP  $x(t)$, which captures many of same features (e.g. the two attractors) as the standard 2D phase-portrait in Figure~\ref{fig:Lorenz}d despite not directly including $y(t)$. Related to this observation, a basic tenet of TLPP is that a single measurement of a sufficiently coupled system contains information about the all state variables when properly time-lagged~\cite{takens1981}.  In preparation for EMD, we convert the TLPP into a probability distribution function (PDF), specifically a 2D histogram, using a $20\times20$ uniform grid as shown in Figure~\ref{fig:Lorenz_TLPP}b.  
		
		The utility of TLPP in this work is the presence of repeatable trajectories (manifolds) in these higher dimensions as they provide a sufficiently unique representation (a ``fingerprint'') of the underlying dynamics.  Please note that these 2D representations are not truly unique in the mathematical sense: the trajectories overlap in the unbinned TLPP and binning obscures individual trajectories.  While we stop at 2D, an advantage of using a higher dimensional TLPP representations is that it typically improves the trajectories uniqueness.  The utility of binning the TLPP, in addition to providing a PDF, is that it better isolates the average trajectory, particularly in the presence of non-repetitive dynamics not captured by the manifold (e.g. measurement noise).

	\subsection{Earth mover's distance (EMD) \label{subsec:emd}}
		
		To create an objective function using TLPPs, we need a method that compares a TLPP trajectory to a reference (desired) TLPP and has a minimum when the TLPPs are identical.  In this work, we use the Earth Mover's Distance (EMD).
		
		The EMD, also called the Wasserstein metric, is a measure of the error (or ``distance'') between two PDFs~\cite{rubner1998}.  While there are other algorithms that perform comparable measurements~\cite{gibbs2002}, the EMD is an ideal choice because it i) provides a positive, scalar value regardless of the dimensionality of the two distributions, ii) guarantees a global minimum at zero only if the two distributions are identical, iii) includes a distance component in its computation which aids in convergence when the distributions do not overlap, and iv) comes prepackaged in the \emph{Python Optimal Transport} library~\cite{flamary2021}.  Intuitively, the EMD can be thought of in the following terms:  if both distributions are piles of dirt, then the EMD provides the minimum cost of moving dirt from the first pile to the second.  Here, cost is the amount (or volume) of dirt to be moved multiplied by the distance moved; this is similar to the the physics concept of ``work''.   Mathematically, the EMD is defined as
		
		\begin{equation} \label{eq:EMD}
			\begin{split}
				\text{EMD}(f, g) = \sum_{i=1}^m \sum_{j=1}^n M_{ij} d_{ij}
			\end{split}
		\end{equation}
		
		\noindent where $f$ and $g$ are the two discretized PDFs, $i$ and $j$ are the multi-dimensional indices of each cell (bin) in $f$ and $g$, respectively, $M_{ij}$ is the value (i.e. histogram bin count) transferred from $f_i$ to $g_j$, and $d_{ij}$ is the Euclidean distance between $f_i$ and $g_j$.  One constraint on this implementation of the EMD is that there must be an equal amount of dirt in each distribution, i.e. $\sum_i f_i=\sum_j g_j$. In this work, this requires that the total number of points ($N$) in each measurement of $x(t)$ be the same for both TLPPs. 
		
\begin{figure}
			\includegraphics[]{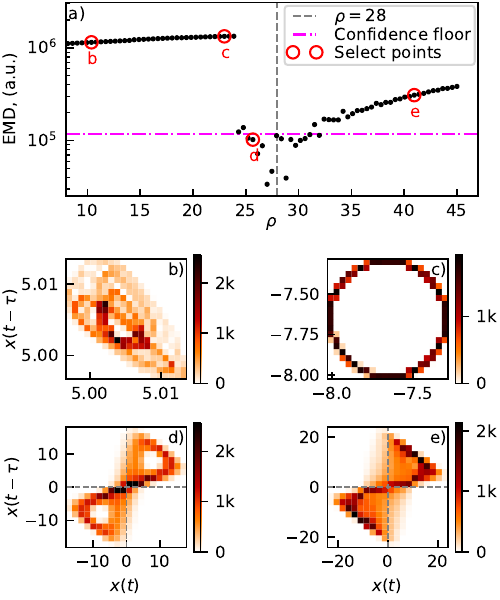}
			\caption{\label{fig:EMD_Lorenz}  a) Shows that EMD, applied to TLPPs across a range of values for $\rho$, provides an objective function that can be minimized.   b-e) Shows the TLPPs at four illustrative values of $\rho$.   }
		\end{figure}

		Figure~\ref{fig:EMD_Lorenz}a shows the EMD measurement between the reference TLPP of the Lorenz system where $\rho=28$ (Figure~\ref{fig:Lorenz_TLPP}b) and the Lorentz system solved across a range of values of $\rho$.   As desired, a minimum occurs around the reference point, $\rho=28$, meaning that the EMD-TLPP objective function can be minimized.  For illustrative purposes, Figures~\ref{fig:EMD_Lorenz}b to~\ref{fig:EMD_Lorenz}e show the 2D TLPP for four points indicated in Figure~\ref{fig:EMD_Lorenz}a, and the calculated EMD value is smaller when the TLPP trajectory is visually similar to the reference trajectory (Figure~\ref{fig:Lorenz_TLPP}b).  
		
		Figure~\ref{fig:EMD_Lorenz}a shows a fair amount of variability in the EMD value near the minimum around $\rho=28$.  This is, in large part, due to the chaotic nature of the Lorenz system.  Measurement noise, while not present here, would produce a similar effect.  This variability limits our confidence in the location of the minimum, which effectively serves as a floor on our confidence.  
		
\begin{figure}
			\includegraphics[]{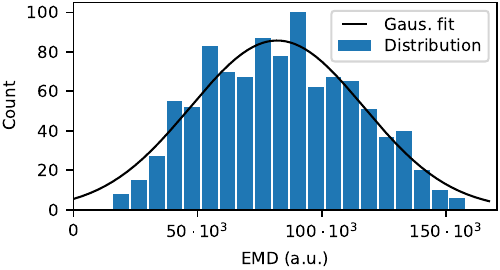}
			\caption{\label{fig:EMD_Lorenz_ICs}  Histogram of the EMD results of the Lorenz system solved 1000 times while perturbing its parameters about its nominal values.  This resulting distribution helps define an effective confidence floor for Figure~\ref{fig:EMD_Lorenz}a.  }
		\end{figure}

		To better characterize this floor, we perform a Monte Carlo uncertainty analysis~\cite{coleman1999} by solving the Lorenz system 1000 times with a $1\%$ perturbation on each of the reference parameters: $\left\{ \sigma \text{, } \rho \text{, } \beta \right\} = \left\{ \mathcal{N}(10, 0.1) \text{, } \mathcal{N}(28, 0.28)  \text{, } \mathcal{N}(8/3, 0.027)  \right\}$. Here, $\mathcal{N}(\mu, s)$ is the normal (Gaussian) distribution with mean ($\mu$) and standard deviation ($s$).  Figure~\ref{fig:EMD_Lorenz_ICs}  shows a histogram of the resulting EMD measurements.   We fit a Gaussian to the distribution, which has a mean of $\mu=82\cdot10^3$ and a standard deviation of $s=35\cdot10^3$.  Therefore, we define our confidence floor as $\mu+s = 117\cdot 10^3=10^{5.07}$, where below this value, the absolute minimum is difficult to discern.  This confidence floor is shown in Figure~\ref{fig:EMD_Lorenz}a.
		Note that adding more measurements and using a statistical model (i.e. Gaussian Process models, described below) improves one's ability to approximate a minimum below this floor.  Note that we similarly perturbed the initial conditions (instead of the parameters), and we found a very similar distribution to Figure~\ref{fig:EMD_Lorenz_ICs}.

	\subsection{Gaussian Process Regression (GPR) and closing the ILC loop\label{subsec:bo}}
		
		The previous subsection established that an EMD measurement between a TLPP and a reference TLPP provides an objective function that is minimized when the system's input parameters are set to values that produce roughly the same dynamics as the reference condition: i.e. when $\left\{ \sigma \text{, } \rho \text{, } \beta \right\} \approx \left\{ 10 \text{, } 28 \text{, } 8/3 \right\}$.  In this work, we use Gaussian Process Regression to minimize the objective function and implement it with the scikit-optimize library~\cite{scikitoptimize2021}.
		
\begin{figure}
			\includegraphics[]{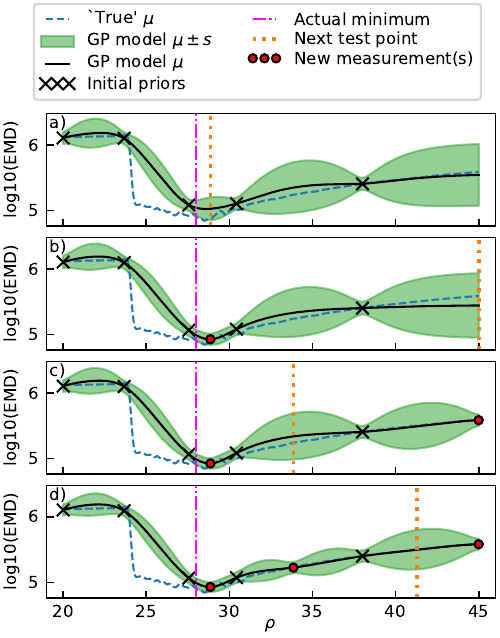}
			\caption{\label{fig:BO_EMD_Lorenz} Four iterations of the ILC control loop applied to the Lorenz system and starting from only five prior measurements.  The GP model is updated after each iteration, and the GPR instructs the system which $\rho$ to test next.  After each iteration, the learned GP model's $\mu$ converges on the ``True'' $\mu$. }
		\end{figure}

		Figure~\ref{fig:BO_EMD_Lorenz} illustrates the GPR process of minimizing the EMD-TLPP objective function over several iterations by controlling $\rho$.  We initialize our model with 5 prior measurements at randomly selected values of $\rho$, where each measurement is assumed to have a non-zero uncertainty.   The GP model is built from a Matern kernel~\cite{rasmussen2005}, where each input parameter (only $\rho$ in this case) is linear-normalized for consistent weighting between parameters.  The EMD result is log10-normalized to better weight the uncertainty across multiple orders-of-magnitude.  Using the uncertainty of the GP model, a negative expected improvement (EI) function predicts the most likely location of the minimum and recommends this as the next point to test.  The EI hyperparameter, $\xi$, provides a balance between exploring ($\xi>1$) and exploitation ($\xi<1$) within the model's parameter space. For this work, we default to $\xi=0.1$, which weights exploitation over exploration.  The resulting mean and standard deviation of the resulting GP model is shown in Figure~\ref{fig:BO_EMD_Lorenz}a.  
		For context, the location of the actual minimum ($\rho=28$) is also plotted.  To be clear, the controller has no knowledge of the actual minimum and instead must rely on its continually updated GP model to find the minimum.
		For additional context, Figure~\ref{fig:BO_EMD_Lorenz} shows a ``True'' GP $\mu$ which is the mean of a GP model calculated from EMD measurements at 1000 uniformly spaced values of $\rho$.  After each iteration, the learned GP model converges on the ``True'' model.   Note that even the ``True'' $\mu$ does not perfectly identify the actual minimum due to high variability near the minimum and is therefore not a perfect representation.
		
		Next, the controller changes $\rho$ to the predicted minimum, the simulation is run again, a measurement is taken, the EMD compares the TLPP of the new measurement with the reference TLPP, the GP model is updated, and EI uses the new GP model to predict the next test point.  The result of one iteration of this loop is shown in		Figure~\ref{fig:BO_EMD_Lorenz}b, and the results of the next two iterations are shown in Figures~\ref{fig:BO_EMD_Lorenz}c and~\ref{fig:BO_EMD_Lorenz}d, respectively. After each iteration, the GP model gains new information which allows it to reduce its overall uncertainty and increase its confidence in the location of the absolute minimum.  This algorithm will run until either a specified level of confidence is reached by the simulation or a maximum number of iterations is reached.

\section{Results: Application to the Lorenz system \label{sec:results}}

	Having described the control loop in Sec.~\ref{sec:architecture}, we next apply it to the Lorenz system for three distinct cases: a single-parameter case, a two-parameter case, and a two-parameter ``robust'' case.  These cases allow us to test the efficacy of our controller as well as identify its key aspects and future work.  
	
	
	\subsection{Single-parameter control case}
		
\begin{figure}
			\includegraphics[]{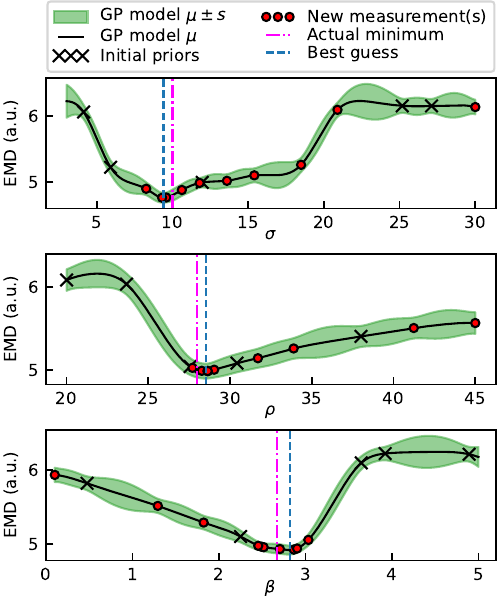}
			\caption{\label{fig:1D_results} Results of  single-parameter control case for $\sigma$, $\rho$, and $\beta$, respectively.  The width of the minima (i.e. the gradient the minima) contains information about the sensitivity of the system's dynamics to its respective parameter. }
		\end{figure}

		In this first case, single-parameter control is applied to each of the three Lorenz parameters ($\sigma \text{, } \rho \text{, } \beta $), independently.  For each, five randomly spaced measurements are used as the initial prior dataset, and the control system performs 10 additional iterations.  Once complete, the controller provides the minimum of the GP model's mean ($\mu$) as its best guess of the minimum.  The result of each test is shown in Figure~\ref{fig:1D_results}, with the actual minimum indicated for context.   
		
		For each parameter, the controller explores the range of values and reduces the model's uncertainty while seeking the minimum.  The final guess of the minimum for each sub-case qualitatively agrees well with the actual minimum.  The width and gradient of the distribution around the minimum carry implications for the sensitivity of the Lorenz system to each individual parameter.  For instance, Figure~\ref{fig:1D_results}a shows that $\sigma$ has the shallowest minimum, which implies that the Lorentz system is less sensitive to $\sigma$ than $\rho$ and $\beta$ at this particular set of coordinates.  This highlights the importance of quantifying the sensitivity of the system's dynamics to its parameters and then optimizing the controller accordingly.
		
	\subsection{Two-parameter control case \label{subsec:2d_results}}
		
\begin{figure}
			\includegraphics[]{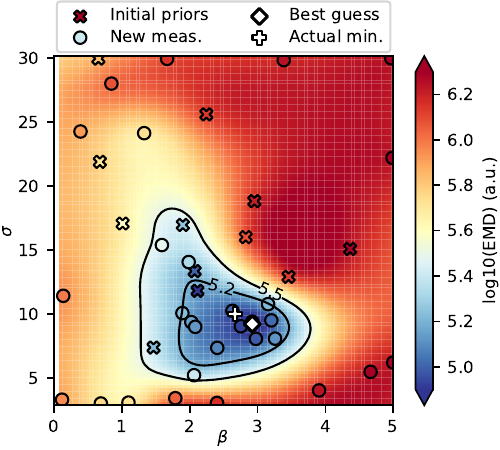}
			\caption{\label{fig:2D_results} Results of the two-parameter control case using $\beta$ and $\sigma$ and fixing $\rho=28$ constant.  The background contour shows the GP model's mean ($\mu$).  }
			\vspace{0.5cm}
			\includegraphics[]{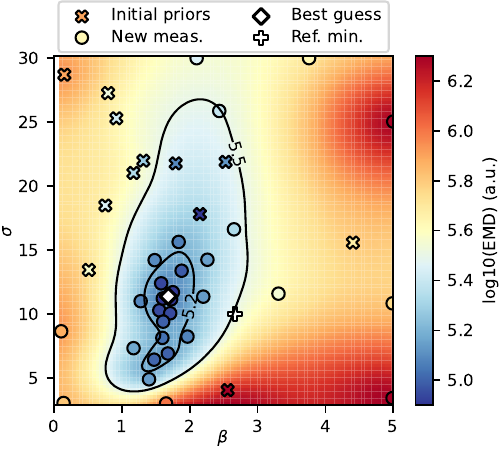}
			\caption{\label{fig:2D_shifted_rho} Results of the two-parameter ``robust'' control case that simulates the robustness of the controller to the abrupt drift of $\rho$ from $\rho=28$ to $\rho=40$.  Despite being unable to measure and control $\rho$, the controller converges on a set of control inputs that are different than what produced the reference condition but still result in similar dynamics.  }
		\end{figure}

		Next, the controller controls two-parameters simultaneously ($\beta$ and $\sigma$) to demonstrate multi-parameter control.  Before closing the loop, the algorithm makes 12 randomly spaced measurements to create the initial prior dataset.  The controller is then run for an additional 30 iterations and the resulting model is shown in Figure~\ref{fig:2D_results}, where the colored contour is the mean of the GP model.   The resulting best guess for the minimum is in reasonable qualitative agreement with the actual minimum for both parameters.  This agreement could be improved by running additional simulations or better tuning the various TLPP and GPR hyperparameters.
		
		Additionally, we have added contour lines at 5.2 and 5.5 (chosen arbitrarily), which could hypothetically represent a threshold for quality or success of convergence.  In other words, the regions within these lines could represent a bounded but continuous range (volume) of input parameters that lead to acceptable dynamics.  This provides a more flexible approach to the discrete parameter mapping discussed in the introduction.

	\subsection{Two-parameter ``robust'' control case \label{subsec:2d_robust_results}}

		Next, we test the system's robustness to missing information and uncontrolled parameters.  To motivate this test, imagine that a system exists and that its dynamics are dependent on a finite number of parameters.  For at least one of these parameters, the controller is either i) not measuring it (i.e. an unmeasured parameter) or ii) is measuring it but has no ability to control it (i.e. an uncontrolled parameter).  Examples in Hall thrusters include internal cathode temperatures, which are difficult to directly measure and drift with time, and environmental background pressure, which changes with spacecraft's altitude and cannot be directly controlled. 
		
		To test this, we repeat the two-parameter ($\sigma$ and $\beta$) control test but simulate an abrupt drift in the unmeasured parameter, $\rho$.  Our reference condition is solved at $\left\{ \sigma \text{, } \rho \text{, } \beta \right\} = \left\{ 10 \text{, } 28 \text{, } 8/3 \right\}$, but we $\rho$ is set to $\rho=40$ when the controller is operating.  This change ensures that the controller has no ability to either measure or change $\rho$ and therefore cannot access or replicate the complete set of parameters that produced the reference condition. 
		
		Figure~\ref{fig:2D_shifted_rho} shows the results of this test.  Despite not having access to $\rho=28$, the controller successfully converges on a solution that provided similar dynamics (e.g. log10(EMD)$<5.2$) by finding an alternate set of control inputs ($\beta=1.73$ and $\sigma=11.1$) that best reproduced the reference condition.  This adaptability is a form of robustness to missing information and uncontrolled parameters.  
		
		There is another way to think about the results is these results.  Specifically, Figure~\ref{fig:2D_shifted_rho} is effectively a 2D parameter map that tracks the level of ``similarity'' between the reference dynamic and the dynamics at each set of inputs (coordinates).  You can also think of this map as a 3D map with coordinates $\sigma$, $\rho$,  and $\beta$, and Figures~\ref{fig:2D_results} and~\ref{fig:2D_shifted_rho} are merely two 2D slices within it at different $\rho$ values.  This raises two points.  
		
		First, the more parameters we include in our GP model, the more predictive the model becomes and the robust it becomes to changing conditions.  This indicates that identifying and measuring all dynamically sensitive parameters (i.e. the entire measurable parameter space) is important (but not necessarily required) for robust control.  
		
		Second, the convergence in Figure~\ref{fig:2D_shifted_rho} occurs because the ``similarity'' region extends throughout the third dimension ($\rho$), albeit at shifted $\sigma$ and $\beta$ coordinates.   If this ``similarity'' region did not extend throughout the third dimension, the convergence would have been much poorer.  This realization establishes a requirement for successful convergence with this method: an adequate minimum must exist within the controllable parameter space.  Therefore, identifying and including all dynamically sensitive parameters as actuators (i.e. the controllable parameter space) is also important for robustness.

\section{Discussion: Transitioning to application and future work \label{sec:discussion}}

	The previous section showed that this controller can control a simple, well-studied toy problem.  Therefore, the next step in the controller's development is applying it to application-relevant simulations and experiments.  In this section, we discuss the future work that will be required to make this transition.
	
	In this work, we provided a ``known'' and ``desirable'' reference trajectory for the controller to reproduce.  However in real-life applications, identifying a desired reference trajectory will likely be non-trivial and also be heavily dependent on the particular application.  One simple approach may be to make the reference trajectory completely stable (i.e. having no oscillations).  Complicating this approach is the fact that system oscillations (instabilities) are frequently unavoidable and even sometimes beneficial.  For example, small-amplitude combustion instabilities in liquid rocket engines can increase combustion efficiency through better propellant mixing~\cite{yang1993}.  Another approach may be to develop a ``safe'' oscillation in laboratory testing and use this as the reference condition; this is a variation on the ``fly-as-you-test" approach to flight operation.  Alternatively, this control system could be used in laboratory testing to reproduce measured flight conditions to better study facility-to-flight reproducibility; this is a variation on ``test-as-you-fly''.  
	
	The control system, diagrammed in Figure~\ref{fig:control_architecture}, is more than a single system, it is potentially a broad architecture capable of controlling fast, nonlinear, oscillatory dynamics across many aerospace applications.  Our choice of TLPP, EMD and GPR in this work is a merely a single implementation (our first pass) of this architecture.  Future work will include identifying other algorithms and variations that may be better suited to particular applications.  
	
	As discussed earlier with our Hall thruster example, there will inevitably be cases where we are neither i) able to measure every parameter we control or ii) control every parameter that we can measure.  In addition, we identified that including every ``important'' measurable and controllable parameter should improve our controller's robustness, and we believe that this ``importance'' metric is related to the system's dynamical sensitivity to each parameter.  Future work will focus on these questions.  
	
	In this work, we introduced GP and TLPP, in part, to provide robustness to non-repetitive dynamics, including noise, but we did not directly include measurement noise as part of this study.  Future work will include noise and other real-world distortions (e.g. imperfect state observers) of measurements.
	
	In discussing this controller, we have discussed (or alluded to) two types of stability, and we clarify each here.  The first is the controller's stability, i.e. the controller's ability to consistently converge on a solution.  EMD and GPR~\cite{berkenkamp2015} both improve this stability but at the cost of computational complexity.  Another possible source of controller instability is when the controller cycle time is faster (i.e. smaller) than $\tau_{slow}$, which results in new measurements, updates, and actuation before the system can converge on a dynamical state.  Guaranteeing and quantifying this stability will be topics of future work.  
	The second stability is the dynamical stability of the system (i.e. are the oscillations in system's state variables growing, steady, or decay.  Future work will focus on improved identification, classification, and avoidance of these instabilities. 
	

\section{Conclusions \label{sec:conclusions}}

	In this work, we developed a control framework  based on Iterative Learning Control (ILC) with the potential to address a broad family of nonlinear, chaotic, and high-frequency dynamics (instabilities) that hinder aerospace vehicle development and deployment.   ILC is unique in that it uses a slower, parameter-tuning approach to control dynamics that are traditionally considered too fast for real-time control, a common feature of aerospace instabilities.  
	In our approach, we have selected three core algorithms (Time-Lagged Phase Portrait, the Earth Mover's Distance, and Gaussian Process Regression), and this particular combination addresses ILC's traditional weakness at capturing non-periodic (nonlinear, chaotic, and noisy) dynamics.  The use of Gaussian Processes additionally allows for online learning as the model improves with each cycle and therefore allows for a higher probability of producing desired dynamics.  
	
	In this work, we demonstrated the controller's effectiveness in controlling the Lorenz system of coupled ordinary-differential-equations, and this resulted in a number of notable findings: i) the system's dynamics are more strongly sensitive to some parameters than others and the controller should be optimized accordingly, ii) the controller is capable of multi-parameter control, iii) the learned GP model provides a bounded but continuous range of input parameters that lead to acceptable dynamics, which potentially provides a more flexible alternative to discrete parameter mapping discussed in the introduction,  and iv) identifying and measuring all dynamically sensitive parameters for both the measurable and controllable parameter spaces is important for robust control. 
	
	While this work introduces a compelling framework for controls of high-speed, aerospace dynamics, additional research is still required.  This includes: i) optimizing the controller's hyperparameters to improve convergence rates, cycle times, and robustness, ii) identifying alternate algorithms with this controller's framework and the advantages that each may provide, iii) applying this control system to laboratory experiments and more application relevant simulations for validation testing and to identify other application-relevant hurdles (e.g. measurement noise), iv) developing metrics for quantifying parameter sensitivity, v) developing methods for identifying previously unrecognized and dynamically important parameters for both measurement and actuation, vi) further improving robustness in the presence of (inevitable) missing information or uncontrollable parameters, and vii) developing strategies for developing ideal and repeatable reference trajectories.

\appendices

	
	\section{The Lorenz system's characteristic frequency \label{app:f_lorenz}}
		
\begin{figure}
			\includegraphics[]{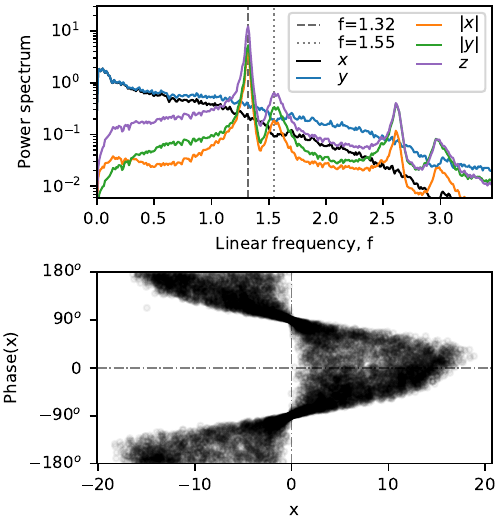}
			\caption{\label{fig:lorenz_fft} a) Shows the power spectrum of the Lorenz system and that it does not capture the oscillation frequency peaks for $x$ and $y$. b) Shows that the phase of $x(t)$ shifts $180^o$ when it transitions between the positive and negative attractors.  This highlights that $x(t)$ and $y(t)$ both are non-periodic, which explains the absence of their peaks in a).   }
		\end{figure}

		Figure~\ref{fig:lorenz_fft}a shows the power spectrum of the Lorenz system (Sec.~\ref{sec:Lorenz}) calculated using Welch's method~\cite{proakis1995} with $100N$ points, a $0.1dt$ step size, a segment length of $N$, and with no segment overlap.  Of the three state variables, only the power spectrum of $z(t)$ shows distinct frequency peaks: a primary peak (the characteristic frequency) at $f=1.32$, a secondary peak at $f=1.55$, and harmonics of these two frequencies.  In contrast, the power spectrums of $x(t)$ and $y(t)$ show only a broad spectrum distribution with no peaks despite having clear oscillations (see Figures~\ref{fig:Lorenz}a and~\ref{fig:Lorenz}b).  This occurs because  $x(t)$ and $y(t)$  are non-periodic.  Specifically, their phase flips (shifts $180^o$) when each transitions between the positive and negative attractors.  This phase dependency is shown in Figure~\ref{fig:lorenz_fft}b for $x(t)$, where the phase is calculated using a Hilbert transform~\cite{proakis1995}.  The power spectrums of $|x(t)|$ and $|y(t)|$, however, show a similar distribution as $z(t)$.  This works because taking the absolute value of $x(t)$ and $y(t)$ rectifies the signals, which effectively re-flips the phase of the negative attractor so that it is consistent with the positive attractor.  This inability of the Fourier analysis to capture non-periodic dynamics makes it poorly suited as a reduced-order (i.e. using as small of a number of bases as possible) representation of dynamics.
		
	\section{TLPP hyperparameter selection \label{app:tau}}

		The TLPP algorithm has two types of hyperparameters: the dimensionally, $E$, and a set of time steps, $\tau=\{\tau_1, \tau_2, \dots\, \tau_{E-1}\}$, between each of the TLPP coordinates.  The dimensionally is the number of coordinates used in the TLPP representation.  A 2D (E=2) TLPP representation has the form, \{$x(t)$, $x(t-\tau_1)$\}, and an E-dimensional representation using only $x(t)$ has the general form of \{$x(t)$, $x(t-\tau_1)$, \ldots,  $x(t-\tau_{E-1})$\}.
		
		
		To determine the optimal $\tau$ values, we use an algorithm called SMI (shadow manifold interpolation)~\cite{eckhardt2019}, which uses the unbinned TLPP of a measurement of a system, e.g. $x(t)$, to reconstruct another measurement of the same system, e.g. $z(t)$.   The quality of the reconstruction is strongly related to the degree of coupling and causality between the measurements, and we argue that the $\tau$ that leads to best quality reconstruction is also the best choice of $\tau$ for the TLPP work discussed in Sec.~\ref{subsec:tlpp}.
		
\begin{figure}
			\includegraphics[]{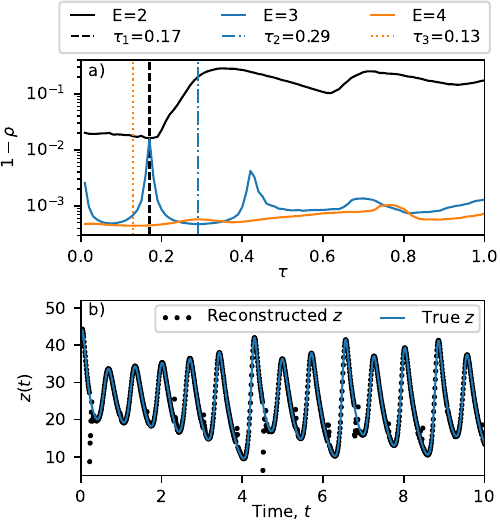}
			\caption{\label{fig:TLPP_parameter_sweep} a) We perform a scan of $\tau$ values for several TLPP dimensions ($E$).  For each $\tau$, we perform an SMI reconstruction and calculate the Pearson correlation, $\rho$,  reconstruction. b) Shows an example $E=2$ SMI reconstruction of $z(t)$ using $\tau_1=0.17$. }
		\end{figure}

		For the Lorenz system, SMI uses $x(t)$ to reconstruct $z(t)$, and the Pearson correlation coefficient~\cite{eckhardt2019}, $\rho$, quantifies the ``goodness of fit'' between the reconstructed $z(t)$ and the simulated $z(t)$.  Note that $\rho \rightarrow 1$ as the fit (positive correlation) improves, and $\rho$ is bounded, $-1\leq \rho \leq 1$.  For the 2D ($E=2$) case, we perform this reconstruction over a range of $\tau_1$ as shown in Figure~\ref{fig:TLPP_parameter_sweep}.  We find that $\tau_1=0.17$ provides the best 2D reconstruction with $\rho=0.984$. For $E=3$, we take a greedy-algorithmic approach (set $\tau_1=0.17$ and scan across $\tau_2$) to find that the $\tau_2=0.29$ provides the best 3D reconstruction with $\rho=0.9995$.  Repeating for $E=4$, and we find $\tau_3=0.13$ with $\rho=0.9996$.  Higher dimensional testing provides marginal improvement.  For this work, we decided that a 2D reconstruction is adequate.

		While we do not have definitive evidence, we suspect that the ideal $\tau_1$ value is related to the characteristic frequency ($f_c$) of the system with the following relationship,
		
		\begin{equation} \label{eq:tau_ideal}
			\begin{split}
				\tau_{1,ideal} \approx \frac{1}{4} \frac{1}{f_c}.
			\end{split}
		\end{equation}  	
		
		\noindent For the Lorenz system, $f_c\approx1.32$ (see Figure~\ref{fig:lorenz_fft}), which results in $\tau_{1,ideal}\approx 0.19$.  This is close to our value of $\tau_1=0.17$.  We believe that the $1/4$ coefficient in Eq.~\ref{eq:tau_ideal} effectively phase shifts the input signal by 90 degrees, and together the two signals \{$x(t)$, $x(t-\tau_{1,ideal})$\}  provide a sine-cosine basis.  This idea is similar to a Hilbert transform~\cite{proakis1995}.

\section*{Acknowledgments}

	This work was supported NRL (Naval Research Laboratory) base funding and also by the Air Force Research Laboratory through Dr. Daniel Eckhardt.

\section*{Data Availability}

	The code that supports the findings of this study are available on Zenodo~\cite{brooks_zenodo_2024}.

\bibliographystyle{IEEEtran}
\bibliography{biblio}

\end{document}